\documentclass[final,5p,times,twocolumn]{elsarticle}

\usepackage{graphicx}
\usepackage{amssymb}
\usepackage{amsthm}
\usepackage{amsmath}
\numberwithin{figure}{section}
\usepackage{lineno}
\usepackage{multirow}
\usepackage[toc,page,title,titletoc,header]{appendix} 

\begin{document}

\begin{frontmatter}


\title{Cystoid macular edema segmentation of Optical Coherence Tomography images using fully convolutional neural networks and fully connected CRFs}


 \author{Fangliang Bai}
 \ead{F.Bai@kent.ac.uk}
 \author{Manuel J. Marques}
\ead{M.J.Marques@kent.ac.uk}
\author{Stuart J. Gibson}
\ead{S.J.Gibson@kent.ac.uk}
 \address{School of Physical Sciences, University of Kent, Canterbury, CT2 7NH, UK.}



\begin{abstract}
In this paper we present a new method for cystoid macular edema (CME) segmentation in retinal Optical Coherence Tomography (OCT) images, using a fully convolutional neural network (FCN) and a fully connected conditional random fields (dense CRFs). As a first step, the framework trains the FCN model to extract features from retinal layers in OCT images, which exhibit CME, and then segments CME regions using the trained model. Thereafter, dense CRFs are used to refine the segmentation according to the edema appearance. We have trained and tested the framework with OCT images from 10 patients with diabetic macular edema (DME). Our experimental results show that fluid and concrete macular edema areas were segmented with good adherence to boundaries. A segmentation accuracy of $0.61\pm 0.21$ (Dice coefficient) was achieved, with respect to the ground truth, which compares favorably with the previous state-of-the-art that used a kernel regression based method ($0.51\pm 0.34$). Our approach is versatile and we believe it can be easily adapted to detect other macular defects.
\end{abstract}


\end{frontmatter}


\section{Introduction}
\label{S:1}
Optical coherence tomography (OCT) has been regarded as one of the most important test instruments in ophthalmic diagnosis \cite{huang1991optical} over the last two decades. During the last decade, the performance of OCT systems experienced a significant improvement, with several variants being introduced and developed further \cite{bradu2015master,marquesa2017polarization}. Spectral-domain OCT (SD-OCT) is a pivotal technique which was first introduced in 2006, paving the way for high-resolution and fast-acquisition OCT systems \cite{de2003improved}. SD-OCT utilizes a broadband optical source and an interferometer, which is interrogated by a high-speed spectrometer. The spectra detected exhibit modulations which are proportional to the depth locations of the features inside the object under study. This method can achieve high at scanning speeds, upwards of 300,000 A-scans per second \cite{potsaid2008ultrahigh}. In ophthalmic applications, SD-OCT is used as an ancillary test for imaging both the anterior and the posterior part of the eye, covering regions such as the cornea, macula, optical nerve, and the choroidal layer \cite{adhi2013optical}. Due to the capabilities of long-range imaging and clear visualization enabled by SD-OCT, it has become straightforward to detect symptoms of age-related macular degeneration (AMD), diabetic macular edema (DME) and central serous chorioretinopathy etc. \cite{manjunath2010choroidal,margolis2009pilot,fung2007optical,haouchine2004diagnosis,yang2013optical}.

Cystoid macular edema (CME) is the major cause of blindness in patients with diabetic retinopathy. This ocular manifestation is becoming a common health issue in ophthalmic disease \cite{klein1998wisconsin}. According to research \cite{aiello1994vascular}, approximately 25\% of patients with diabetes suffer from diabetic retinopathy and the likelihood of developing CME increases the longer a patient lives with diabetes. The morphology of diabetic retinopathy has four classifications; diabetic macular edema (DME), retinal detachment, diffuse retinal thickening, and vitreomacular interface abnormalities \cite{otani1999patterns}. Apart from diabetic retinopathy, CME is also a cause of visual degradation in other pathological conditions, including AMD, vein occlusion, epiretinal membrane traction, etc. \cite{johnson2009etiology}. Since OCT is a non-invasive imaging method which allows the detection of CME in a reasonably simple manner, many clinicians prefer the use of OCT for the assessment of patients with diabetic retinopathy \cite{kozak2008discrepancy,hee1995quantitative}. The advantage of using OCT is that it eases the quantitative assessment, thus presenting more structural information than the qualitative evaluation performed with fundus photography. Therefore, measuring the quantity of CME and monitoring its change over time is of significant importance in clinical assessment.

Segmentation of CME in OCT images is always a challenging task for clinicians. OCT systems with high acquisition rates yield large amounts of 3D data, consisting of hundreds of B-scan images in each volume. Manual annotation of each B-scan image would require significant processing time which is impractical for clinical applications. Furthermore, when carrying out the quantitative assessment it is preferable to have the segmentation in a 3D format, since this provides more morphological information. However, the imaging acquisition process presents deviations due to involuntary movements of eyes and head of the subject. These movements affect continuity and smoothness within interval frames of the OCT volumes. This defect also introduces difficulties in manual annotation. Apart from the difficulty mentioned above, a severe retinopathy can result in a high diffusion of boundaries between healthy and diseased tissue. This diffusion is likely to cause inconsistencies of manual annotation results between different experts. The manual segmentation of CME areas can therefore become very subjective. Hence, various automated segmentation approaches have been developed to produce time-efficient and stable detections yielding different parameters, such as corneal angle, retinal layer boundaries, layer thickness and CME area \cite{dufour2013graph,chiu2015kernel,srinivasan2014automatic,chiu2010automatic,kafieh2013intra}.

In this article, we propose an automated method to segment regions, exhibiting polymorphous cystoid macular edema, in retinal OCT images using a deep-learning based approach which has been used extensively in natural image tasks \cite{MartinFTM01}. To the best of our knowledge, our work is the first to employ fully convolutional neural networks with dense conditional random fields in segmentation tasks involving retinal OCT imagery.

The article is structured as following. In Section \ref{S:2}, the CME manifestation , the existing methods, the fully convolutional neural network and the fully connected CRFs are reviewed separately. In Section \ref{S:3}, we present the framework of CME segmentation on retinal OCT images. In Section \ref{S:4}, we present and discuss the results. And in Section \ref{S:5} we end the article with conclusion and future work. 

\section{Review}
\label{S:2}
In this section, we briefly review the properties of CME present in OCT images from subjects with diabetic retinopathy, followed by a discussion of the existing methods for CME segmentation, the fully convolutional neural network (FCN) model and the fully connected CRFs (Dense CRFs) procedure as a post-process of refining segmentation.

\subsection{Character of Human CME in OCT image}
\begin{figure}[h]
\label{Fig:1}
\centering\includegraphics[width=1\linewidth]{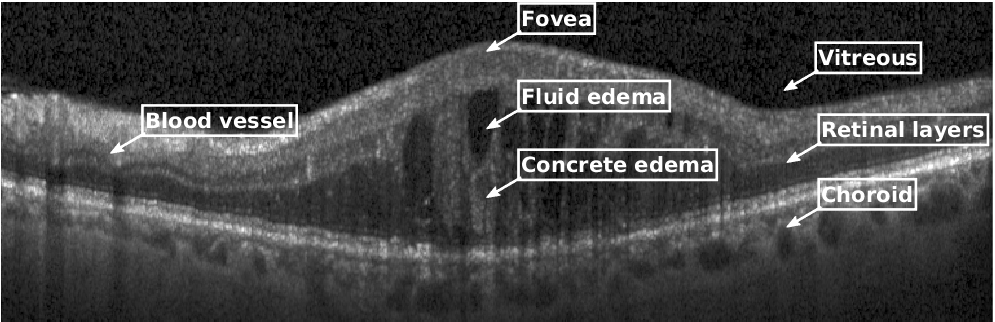}
\caption{A retinal OCT image with DME symptoms in the foveal region from our experimental dataset. The retinal layers are located between the vitreous and choroid layers. The highly reflective dots with shadows underneath are blood vessels and their shadows. In the fovea, retinal layers are detached by a mixture of fluid and concrete edema.}
\end{figure}

\begin{figure}[h]
\label{Fig:2}
\centering\includegraphics[width=1\linewidth]{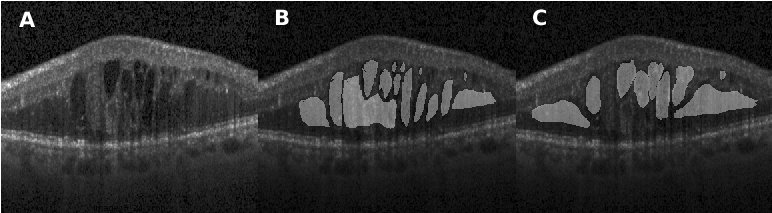}
\caption{Manual segmentation of DME in OCT retinal images carried out by two different graders: (A) the original image; (B) and (C) manual segmentation performed by two different graders. There is a significant difference between the two manually segmented images. The central region, whose appearance is closer to a concrete edema, was annotated by the first grader (B) but not by the second one (C).
This variability may be caused by the OCT image quality, the appearance of the edema and the graders’ experience.}
\end{figure}

An example of an OCT retinal image exhibiting DME is shown in Fig. \ref{Fig:1}. The retinal layers are deformed and discrete with large cystic fluid areas embedded within the sub-retinal layers. The diabetic edema can be categorized into types: fluid and concrete. The fluid regions, embedded in various shapes and sizes, contain sparse tissue that exhibits low reflectivity. Therefore the fluid regions in the OCT images are generally shown as dark blocks surrounded by layers with high reflectivity. Conversely, the concrete regions are formed by denser tissue. The concrete area is not easily distinguishable from healthy tissue in these images, resulting in the boundary between healthy and diseased tissue not being clearly shown and segmented. Conventionally, the medical professionals segment the concrete area by subjectively estimating the thickness and delineating the boundary at what might be a reasonable position based on their experience. Therefore, different professionals may produce inconsistent results. Figure \ref{Fig:2} depicts annotations obtained from two graders (both medical professionals). For example, the delineation of edema does not have a good agreement between the two manually segmented images (B and C). At the end of this article, we will show that the proposed method can objectively outperform manual segmentation in terms of consistency metrics.

\subsection{Existing approaches for CME segmentation}
With the development of OCT imaging systems, the task of segmenting retinal structures has been deeply researched and implemented with various approaches. However, most existing segmentation methods focus on delineating retinal layer boundaries in healthy eyes. One branch of image segmentation methods are based on energy minimization techniques \cite{boykov2006graph, kolmogorov2004energy}. This group of algorithms represents the OCT image as an adjacency graph and each pixel as the node of the graph. The edge of the graph represents the degree of similarity between two connected nodes. Then graph cut theory \cite{ford1956maximal} is used to minimize the summation of the weights of graph edges with the aim of finding the boundaries. A well developed method was proposed by Chiu \textit{et al.} \cite{chiu2010automatic} who used graph searching and dynamic programming to find the shortest path between left and right edges of the graph. This method and its variants explore the characteristics of the retinal structure and calculate the graph path based on the image gradient. This method can produce a decent segmentation result on normal retinal structures. However, the segmentation accuracy decreases when performing the segmentation on pathological retinal structures, where the various morphological conditions which deform the structures induce complexity and lead to poorly delineated boundaries.

Some approaches toward edema fluid segmentation have been implemented which are derivatives of methods which segment layer boundaries. A fluid-filled region segmentation method by Fernandez \cite{fernandez2005delineating} adapted an active contour model to segment individual fluid edema regions. The active contour model is well designed to delineate the retinal lesion boundaries. It employs assumptions based upon a set of initial boundaries. It requires the region to be closed and considers the region to be homogeneous when compared to the surrounding area. More recent work, developed by Chiu \textit{et al.}\cite{chiu2015kernel}, introduced kernel regression to a previously described graph searching framework to classify the layer boundaries and edema regions. In their proposed method, they adapted an iterative Gaussian steering kernel to a supervised classification function that uses local features to classify pixels into a range of categories. Graph theory and a dynamic programming algorithm was used as a classification estimate to segment the retinal layer and the edema regions. In the results they reported, this method demonstrated good accuracy for layer segmentation but the accuracy for edema segmentation was relatively poor.

All the methods reviewed above, include some ad hoc rules built into the mathematical models which are based on the nature of the target object. When the character of the target object becomes varied and complex, more conditions should be considered to maintain the accuracy, which results in the model becoming complicated, and less efficient. Additionally, with more specifically designed rules, the application scope of the approach is narrowed, making it not applicable to multi-purpose diagnosis. In Section \ref{S:4}, we show that our approach can adapt to various conditions and characteristics of the target object. We compared our segmentation results with the results of Chiu \textit{et al.}\cite{chiu2015kernel} which indicates an improvement on edema region segmentation.

\subsection{Fully convolutional neural network model}
\begin{figure*}[h]
\label{Fig:3}
\centering\includegraphics[width=1\linewidth]{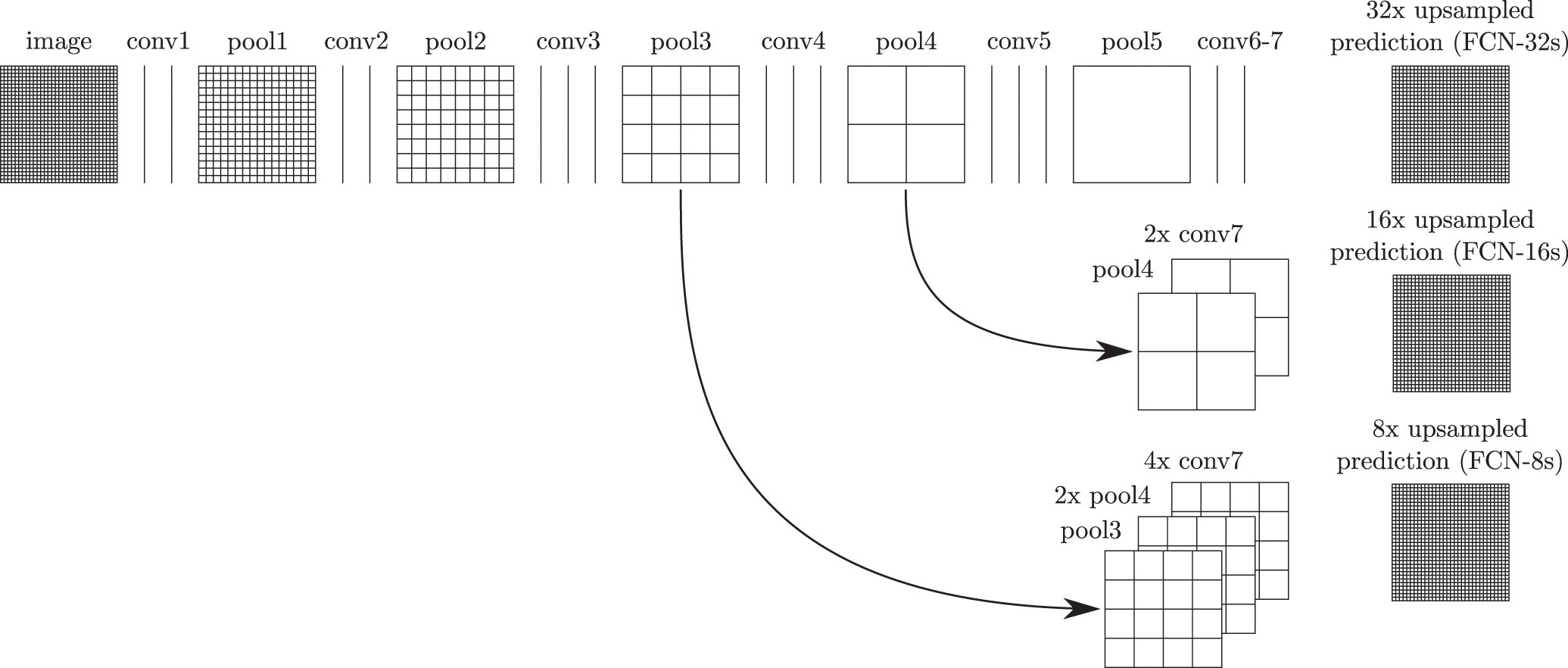}
\caption{The general architecture of a fully convolutional neural network\cite{long2015fully}. The convolutional layers with different filter size carry out the convolution operations with the output of the previous layer or the initial image input. The convolutional layers are followed by the pooling layer that down-samples the dimension of data while mitigating the computational overhead. The fusion link between the network output and that of previous convolutional layers fuses the information of both deep and shallow layers, which refines the spatial segmentation of the image.}
\end{figure*}

A fully convolutional neural network (FCN) \cite{long2015fully} is an end-to-end training model that uses a sequence of convolutional layers to extract and predict the class of pixels of the input image. A common FCN model can be constructed by replacing a fully connected layer of classification networks, such as VGG net \cite{simonyan2014very} and GoogleLeNet \cite{szegedy2015going} , with a convolutional layer and using fine-tuning to exploit the learned convolutional kernel of the classification networks for the segmentation task. The key point of FCN is its skip architecture \cite{bishop2006pattern} that combines detailed information extracted from shallow convolutional layers with semantic information from deep convolutional layers. A common FCN model is formed with a set of convolutional layers, pooling layers, transposed convolutional layers and skip links from pooling layers to transposed convolutional layers. The general architecture FCN \cite{long2015fully} is schematically represented in Fig. \ref{Fig:3}.

Given a two dimensional input image, $I$, of size $M \times N \times 1$, the first convolutional layer convolves the image with $F_{1}$ spatial kernels, each of which have a size of $m_{1}\times n_{1} \times 1$. The convolution for each kernel results in a feature map whose size depends on kernel size and stride size. At the output of the first convolutional layer, the $F_{1}$ feature maps are stacked as an input to the next layer. Multiple kernel convolutions can extract diverse features such as appearance information from the image. However, these also involve dense computations when carrying on into deep convolutional layers. To mitigate the high computational cost, the model inserts a pooling layer after each convolutional layer. The pooling layer reduces the dimension of feature maps by fusing (maximum or average) local spatial information with the kernel \cite{liskowski2016segmenting}. After these operations are carried out in the deep layers, the size of each feature map decreases. The last layer of FCN is an up-sampling layer that operates a transposed convolution (a.k.a deconvolution in Caffe \cite{jia2014caffe}) on the feature maps. The up-sampled feature maps are then fused to generate a heat map that indicates the location of target object. The predicted results may be coarse due to the high down-sampling rate through the network. This can be improved by using the skip link which fuses the predictions from previous pooling layers with the prediction from the final convolution layer. For instance in Fig. \ref{Fig:3}, a $1 \times 1$ convolution layer is inserted after $pool4$ to produce additional class predictions. This prediction is fused with the predictions from $conv7$ by adding a $2\times$ up-sampling layer and summing both predictions. This up-sampling layer is initialized to be a bilinear interpolation of $conv7$, however the parameters are learnable during training. It was found that this strategy improves the prediction by fusing back to the $pool3$. 

During the training process, the weights for convolutional layers can be those of the pre-trained model of modified classification networks rather than just randomly initialized. Then the transfer learning of the neural network can be applied to transferring learned features of the pre-trained network to learn features in the new problem by fine-tuning the parameters during training. The training error between the predicted results and the previously labeled images is calculated after each training batch. Then a back-propagation process \cite{lecun1990handwritten} is utilized to fine-tune the layer weights.

\subsection{Fully connected CRFs classification}
For a given image and corresponding label prediction which could be the heat map produced with the FCN model, the fully connected CRFs model finely corrects the class for each pixel based on feature vectors of image and heat map (the vector contains pixel location and value in different channels) and generates an amended prediction map as the final output. Considering an image $I$ of size $N$ and random field $X$ over $K$ possible pixel-level labels $L = \{l_{1},l_{2},\cdots,l_{k}\}$, a conditional random field $(I,X)$ is characterized by a Gibbs distribution $P(X\mid I) = \frac{1}{Z(I))}\exp(-\sum_{c\in C_{G}} \psi_{c}(X\mid I))$， where $G = (V,E)$ is a graph on $X$ in which each clique $c$ in a set of cliques $C_{G}$ induces a potential $\psi_{c}$. In the fully connected CRFs model, $G$ is a fully connected graph of $X$ and $C_{G}$ is the set of all unary and pairwise cliques which forms the Gibbs energy function of labeling $x\in L^{N}$ in Equation \ref{eq:GibbsEnergy},

\begin{eqnarray}
\label{eq:GibbsEnergy}
E(X) &=& \sum_{c\in C_{G}} \psi_{c}(x_{c})  \nonumber \\ 
&=& \sum_{i}\psi_{u}(x_{i})+\sum_{i<j}\psi_{p}(x_{i},x_{j})
\end{eqnarray}

where $i$ and $j$ range from 1 to $N$. The unary potential $\psi_{u}(x_{i})$ is calculated for each pixel by the FCN model, producing a heat map with label assignment $x_{i}$. The pairwise potential is described by:
\begin{equation}
\label{eq:pairwise}
\psi_{p}(x_{i},x_{j}) = \mu (x_{i},x_{j})\sum_{m=1}^{K}\omega^{(m)}k^{(m)}(f_{i},f_{j})
\end{equation}
where each $k^{(m)}$ is a set of Gaussian kernel functions for the feature vector $f$ of pixel $i$ and $j$ separately, $\omega^{(m)}$ describes the linear combination weights and $\mu$ is a label compatibility function which introduces a penalty for neighboring similar pixels which were assigned different labels. The pairwise function classifies pixels with similar feature vectors into the same class so that different classes are prone to be divided at content boundaries. Then the fully connected CRFs model infers the final label of each pixel by using mean field approximation, which is described in detail in Koltun \cite{koltun2011efficient}.

\section{Method}
\label{S:3}
\subsection{Problem statement}
Given a retinal OCT image $I_{M\times N}$, the goal is to assign each pixel $\{x_{i,j}, i\in M, j\in N\}$ to a class $c$ where $c$ is in  the range of the class space $C = \{c\} = \{1, \cdots , K\}$ for $K$ classes. We consider a 2-class problem that treats the CME regions as the target class and others as background class.

\subsection{Pre-processing}
The Pre-processing of retinal OCT images involves 3D de-noising and region cropping. Due to the speckle noise present in the images, the BM3D algorithm \cite{dabov2009bm3d} is used on each 3D OCT volume. This algorithm is a block-matching de-noising method using sparse 3D transform-domain collaborative filtering, with the aim of generating high contrast OCT images. The de-noising process suppresses OCT speckle noise while sharpening the image, ultimately improving the presentation of CME region. After the de-noising step, the retinal region is cropped from the OCT volume to eliminate the unrelated vitreous objects and the tissues beneath the Choroidal layer.

\subsection{Network architecture of FCN model}
The FCN-8 model adapted the architecture from the VGG-16 network \cite{simonyan2014very}. This architecture fuses the final convolutional layer with the output from the last two pooling layers to make an 8 pixel stride prediction net. We modified the network by adding an additional score layer to reduce the prediction to 2 classes. We also used a dice layer to compute the overlap with a dice similarity coefficient \cite{milletari2016v} between predicted results and ground truth labels. By using the pre-trained weights of the convolutional kernel, we can transfer pre-trained weights to learn new features of retinal OCT images. 

\subsection{Data augmentation}
The original data set only contained 110 OCT images. We enriched the data by applying spatial translations to each image, which involved random translation, rotation, flipping and cropping. The total number of training images was increased to 2800 in this manner, and 20\% of these images were used for validation.

\subsection{Loss function and optimization}
The network is jointly trained with logistic loss and Dice loss. The multinomial logistic loss, used in original FCN model \cite{long2015fully}, provides a probabilistic similarity at the pixel level, comparing the ground truth label and the prediction results. In our model, the loss function was implemented using softmax with loss layer, built in Caffe \cite{jia2014caffe}. We also use Dice loss \cite{milletari2016v} to compute the spatial overlap between prediction and ground truth. During the training process, we have used the stochastic gradient descent (SGD) as an optimizer to reduce the loss function.

\section{Experiments}
\label{S:4}
\subsection{Experimental datasets}
To validate the effectiveness of our approach, the method was tested on the same dataset used in previous work by Chiu \textit{et al.} \cite{chiu2015kernel}. The dataset comprises 110 SD-OCT B-scan images which are collected from 10 anonymous patients with DME. Selected B-scans are located at the fovea and at two opposite sides of the foveal region (1 B-scan at central fovea and 5 B-scans at 2, 5, 10, 15 and 20 slices away from the central region on each side). These B-scan images are annotated for 7 retinal layers and intra-layer fluid sections. The image size is $512\times 740$. For each patient, 11 B-scans of the 3D volume data are annotated by two professional graders using DOCTRAP semi-automatic segmentation software (Duke University, Durham, NC, USA) \cite{lee2013fully} or manually annotated. Specifically, the fluid CME section was firstly annotated using DOCTRAP software package, then reviewed and manually corrected to refine the segmentation. For the concrete regions, the grader performed the annotation from experience by estimating the region in relation to the thickness of the retinal layer.

\subsection{Experimental setting}
In our experiments, we adapted the pre-trained FCN-8 model from Hariharan \textit{et al.} \cite{BharathICCV2011}. The architecture of the network is shown in Table \ref{Tab:1}. The model weights were pre-tuned rather than randomly initialized. A further fine-tuning was performed to fit the OCT images. The hyperparameter settings for the training are as follows: the number of training epochs is 30; the batch size is set to 1; the base learning rate is $1.0e10^{-4}$ and is reduced by an order of magnitude after every 10 epochs; and the validation interval was done after each epoch. For the input batch, we keep the original pixel values rather than those subtracted by the mean value. The experiments were run on a desktop computer with an Intel CPU, 16 GB of RAM, and a NVIDIA 1080Ti GPU with 16GB VRAM.

\begin{table*}[h!]
\centering
\caption{The architecture of the FCN-8 model used in the experiment.}
\label{Tab:1}
\begin{tabular}{cccccccc}
\hline
 & Type & Filter size & Stride & Filter number & Padding & Number of blocks & Fusion Link \\ \hline
\multirow{2}{*}{Block 1} & Convolution & 3 & 1 & 64 & 100 & \multirow{2}{*}{$\times$ 2} &  \\
 & ReLU & - & - & - & - &  &  \\
Block 2 & Max Pool & 2 & 2 & - & 0 & $\times$ 1 &  \\
Block 3 & Convolution & 3 & 1 & 128 & 1 & \multirow{2}{*}{$\times$ 2} &  \\
 & ReLU & - & - & - & - &  &  \\
Block 4 & Max Pool & 2 & 2 & - & 0 & $\times$ 1 &  \\
Block 5 & Convolution & 3 & 1 & 256 & 1 & \multirow{2}{*}{$\times$ 3} &  \\
 & ReLU & - & - & - & - &  &  \\
Block 6 & Max Pool & 2 & 2 & - & 0 & $\times$ 1 &  \\
Block 7 & Convolution & 3 & 1 & 512 & 1 & \multirow{2}{*}{$\times$ 3} &  \\
 & ReLU & - & - & - & - &  &  \\
Block 8 & Max Pool & 2 & 2 & - & 0 & $\times$ 1 &  \\
Block 9 & Convolution & 3 & 1 & 512 & 1 & \multirow{2}{*}{$\times$ 3} &  \\
 & ReLU & - & - & - & - &  &  \\
Block 10 & Max Pool & 2 & 2 & - & 0 & $\times$ 1 &  \\
Block 11 & Convolution & 7 & 1 & 4096 & 0 & \multirow{2}{*}{$\times$ 2} &  \\
 & ReLU & - & - & - & - &  &  \\
Block 12 & Convolution & 1 & 1 & 21 & 0 & $\times$ 1 &  \\
Block 13 & Deconvolution & 4 & 2 & 21 & - &  &  \\ \hline
Block 14 & Convolution & 1 & 1 & 21 & 0 & \multirow{4}{*}{$\times$ 1} & \multirow{4}{*}{$\times$2 fusion} \\
 & Crop & - & - & - & - &  &  \\
 & Element-wise Fuse & - & - & - & - &  &  \\
 & Deconvolution & 4 & 2 & 21 & - &  &  \\
Block 15 & Convolution & 1 & 1 & 21 & 0 & \multirow{4}{*}{$\times$ 1} & \multirow{4}{*}{$\times$4 fusion} \\
 & Crop & - & - & - & - &  &  \\
\multicolumn{1}{l}{} & Element-wise Fuse & - & - & - & - &  &  \\
\multicolumn{1}{l}{} & Deconvolution & 16 & 8 & 21 & - &  &  \\
\multicolumn{1}{l}{Block 16} & Crop & - & - & - & - & $\times$ 1 &  \\
\multicolumn{1}{l}{Block 17} & Convolution & 1 & 1 & 2 & 0 & $\times$ 1 &  \\
\multicolumn{1}{l}{Block 17} & Dice & - & - & - & - & $\times$ 1 &  \\
\multicolumn{1}{l}{Block 19} & SoftmaxWithLoss & - & - & - & - & $\times$ 1 &  \\ \hline
\end{tabular}
\end{table*}

The grayscale OCT image data was reformatted to RGB with channel conversion as required by the network architecture. The training data was stored using an lightning memory-mapped database backend with PNG compression format for feature encoding and label encoding. 20\% of data was used for validation process.

We also modified the parameter setting of the fully connected CRFs model. The original model \cite{koltun2011efficient} was devised for RGB image inference. In order to adapt the model for grayscale images, the parameters of the kernel were changed in the pairwise function. The standard deviation is 2 for the color independent function, and 0.01 for the color dependent term. We set the ground truth certainty to 0.6 in order to obtain the best results.

\subsection{Evaluation metrics}
In order to compare our method with Chiu's work, the evaluation of the framework is done based on same the two metrics reported in \cite{chiu2015kernel}. With a manual segmentation by a grader as ground truth, we measured the Dice overlap coefficient \cite{sorensen1948method} for the fluid segmentation. The Dice coefficient, defined in Equation \ref{equ:dice}, calculates the index of overlay of auto-segmented results $X_{auto}$ and manual annotation $X_{manual}$. Furthermore, we calculated the mean Dice coefficient for each patient and performed the Wilcoxon matched-pairs test across all patients to evaluate the difference between framework results and the corresponding manual results.

\begin{equation}
\label{equ:dice}
\textup{Dice}=\frac{2\left | X_{auto}\cap X_{manual} \right |}{\left | X_{auto} \right |+ \left | X_{manual} \right |}
\end{equation}

\subsection{Segmentation results}
After training the FCN network, we tested the framework with a separate test data set which was not previously seen by the network. The selected data set comprises a set of OCT images exhibiting various morphologies of DME symptoms. In this section we show the segmentation results of the framework compared to manual segmentation results for different DME conditions. We also present the comparison between manual annotations and the framework results where small edema regions failed to be segmented. Throughout the whole study, our results were compared with those from \cite{chiu2015kernel} in both a qualitative and quantitative way using the same metric to show an improvement in terms of segmentation accuracy. 

\begin{figure}[h]
\centering\includegraphics[width=1\linewidth]{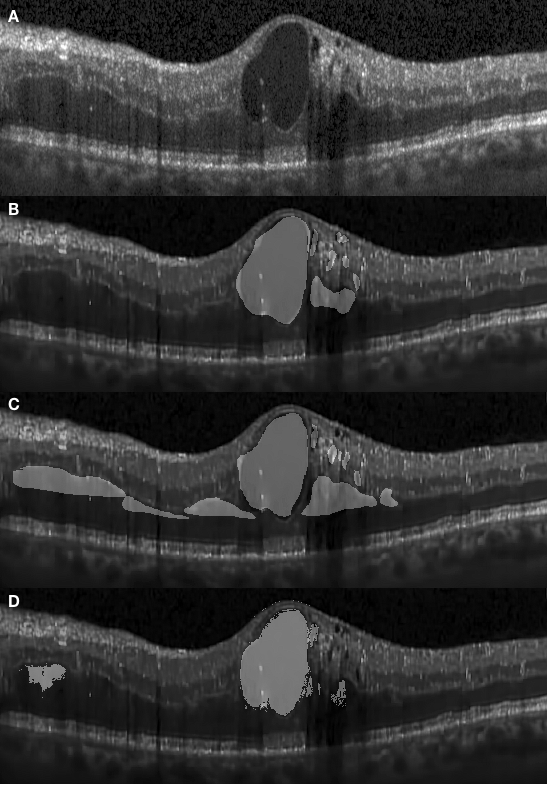}
\caption{The segmentation results of fluid edema by graders and the framework. (A): original image of the fluid edema region. (B) and (C): segmented results by two separate graders. The segmentation of the fluid region is in good agreement, however the region on the left-hand side was not segmented by the first grader. (D): segmented results obtained with the framework. The fluid sections are segmented from the boundary. On the left-hand side, a small region of concrete edema is also detected. However unlike the image obtained from the second grader, the entire region is not covered.}
\label{Fig:4}
\end{figure}

The appearance of fluid edema in OCT images is often shown with a higher contrast against surrounding tissue. Figure \ref{Fig:4} (A) depicts the foveal region affected with a fluid edema. The dominant edema detaches retinal layers and small edema regions surround it sporadically. It can be seen on the framework results that the boundaries of the dominant edema are accurately segmented out, and the smaller edema on the right-hand side is also segmented individually with clear boundaries. Although two graders annotated the same fluid regions, the framework shows an improvement in terms of boundary recall. A significant inconsistency among manual segmentations on the left side of the fovea has also been observed. Because our network was trained using labels previously provided by the first grader, it is prone to be more similar to their results.

\begin{figure}[h!]
\centering\includegraphics[width=1\linewidth]{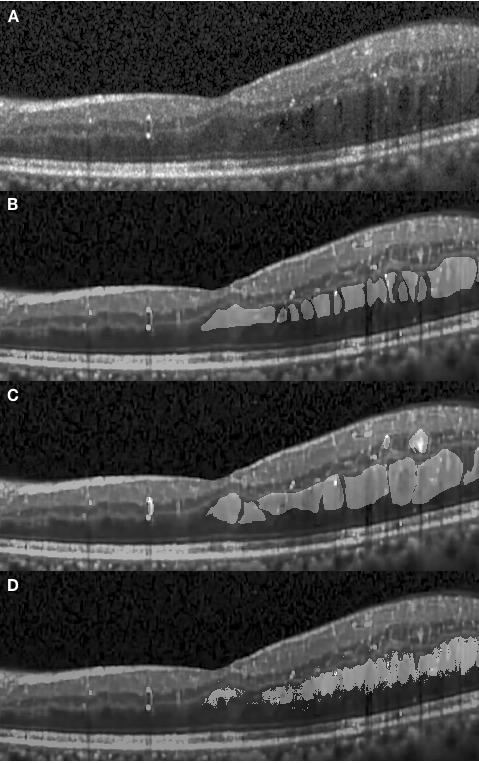}
\caption{The segmentation results of concrete edema by graders and the framework. (A): original image. (B) and (C): segmented results by two separate graders. The segmentation results by two graders are similar. (D): segmented results obtained with our framework. On the right-hand side, most of the region is segmented continuously. The segmented region just under the foveal pit partially covers the small region of concrete edema at the fovea.}
\label{Fig:5}
\end{figure}

Figure \ref{Fig:5} (A) shows a concrete edema embedded within retinal layers on the right-hand side of the fovea region. The distinctive features of this edema are not visually recognizable in this OCT image due to low contrast within the retinal layers. However the presence of the edema causes a deformation of the retinal layers on the right-hand side, with a corresponding change of the layer interval, making them noticeably thicker than those on left-hand side. It is observed in Fig. \ref{Fig:5} (D) that the segmented results obtained with our approach presents a high correspondence to the manually segmented results. However, it is worth noting that our approach has segmented several regions which had been represented separately under manual annotation as a single large one.

\begin{figure}[h!]
\centering\includegraphics[width=1\linewidth]{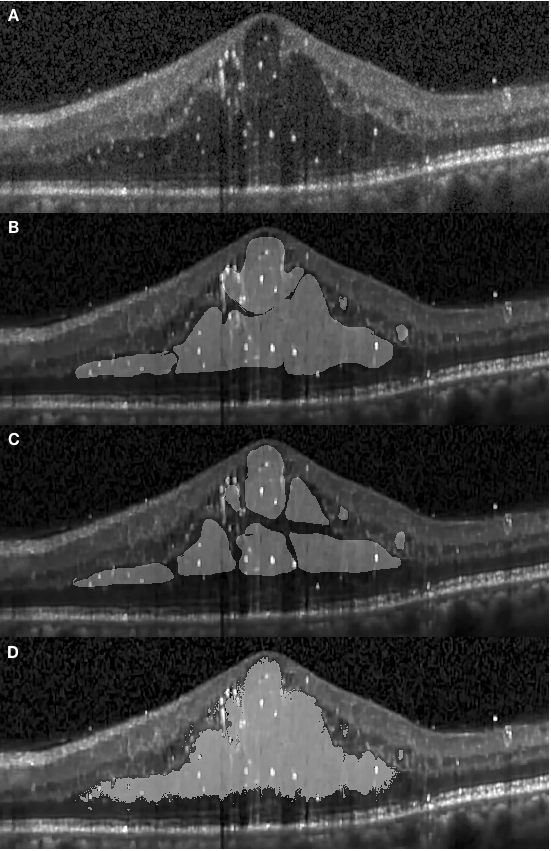}
\caption{Segmentation results for more severe edema as carried out by graders and the framework. (A): original image. (B) and (C): segmented results by two graders. (D): segmented results by the framework. It is seen that both the concrete an fluid regions exhibiting complex appearances are segmented.}
\label{Fig:6}
\end{figure}

\begin{figure*}[h!]
\centering\includegraphics[width=.32\linewidth]{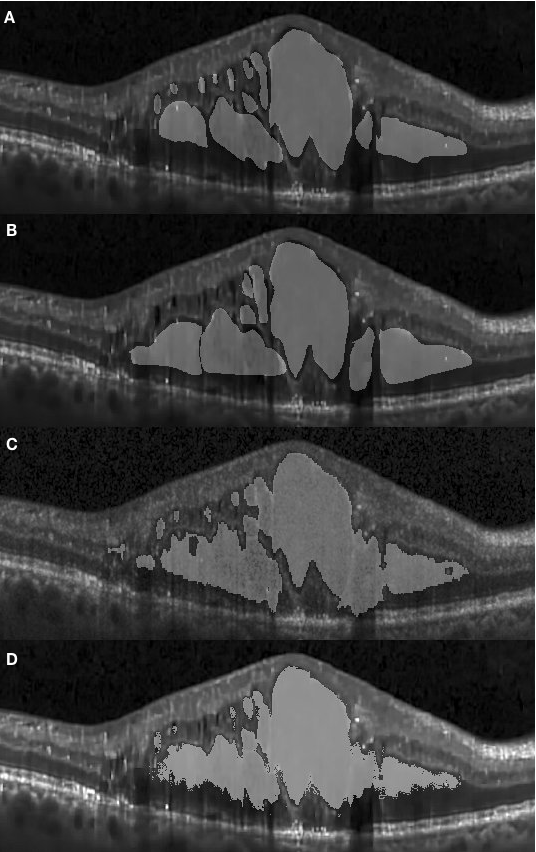}
\includegraphics[width=.32\linewidth]{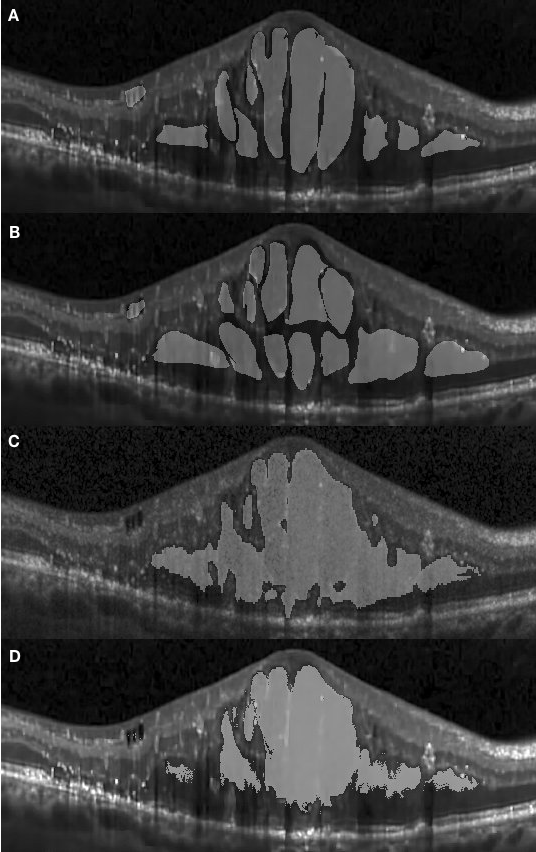}
\includegraphics[width=.32\linewidth]{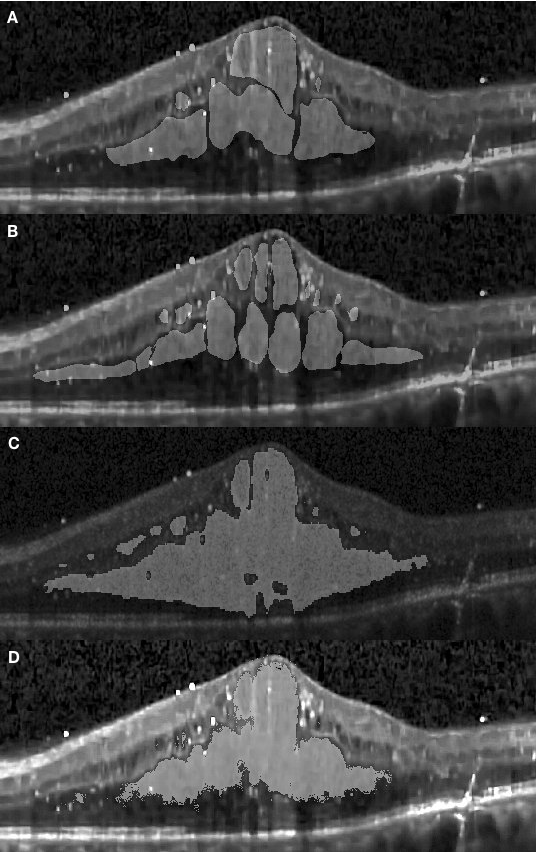}
\caption{A comparison of Chiu's work and our results. (A) and (B): annotations by two graders. (C): Segmentation results from Chiu's work. (D): our results. From the comparison, it can be seen that the results at third row is slightly over-segmented and not coherent to the boundaries. This incoherence is not present in our results.}
\label{Fig:8}
\end{figure*}

In Fig. \ref{Fig:6} (D) we demonstrate results of our framework operating on images depicting more severe morphological conditions. The retinal layer in Fig. \ref{Fig:6} (A) was destroyed by a large section of edema which mixed various shapes of fluid and concrete edemas. The central fluid section is fused with the concrete part on both the left and the right-hand side of the images, where both sections were successfully segmented by our approach. It can be seen that our approach shows that the segmentation achieved is consistent with the appearance of the pathology and accurately covers the entire region. Several small sections were also well segmented but some slimmer region is not detected since it is too slim to be perceived by the network. In Fig. \ref{Fig:8}, we present additional results from severe conditions that present complex structure appearances.

\begin{figure}[h]
\centering\includegraphics[width=1\linewidth]{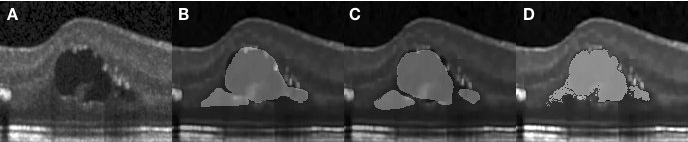}
\caption{The failed case of segmentation result. (A): original image. (B) and (C): segmented results by two graders. (D): segmented results by the framework. It can be seen that part of the concrete edema is not completely segmented where it appears close to the surrounding tissue.}
\label{Fig:7}
\end{figure}

Apart from the segmentation of large areas, there are still some cases where our approach failed to segment the target region accurately. One such event occurred when small target regions were not segmented out, as shown in Fig. \ref{Fig:4} (D). In this case, the small regions were affected by speckle noise and smoothing effects caused by the pre-processing resulting in the small regions being undetectable by the framework, due to their small initial size leading to them being smoothed out after several pooling layers. In another case, some target regions were only partially segmented, as shown in Fig. \ref{Fig:7}.

\begin{figure*}[h!]
\centering\includegraphics[width=.32\linewidth]{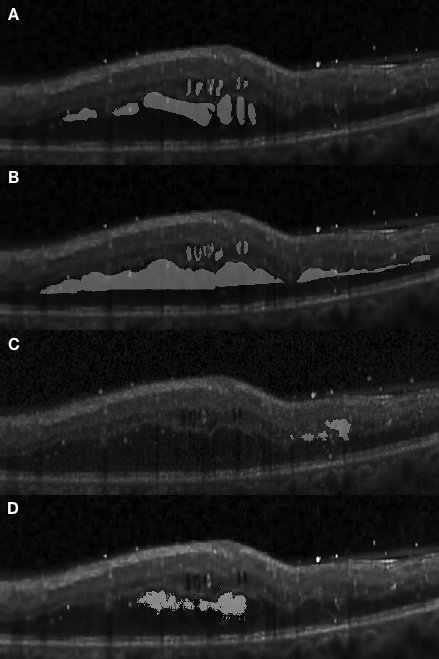}
\centering\includegraphics[width=.32\linewidth]{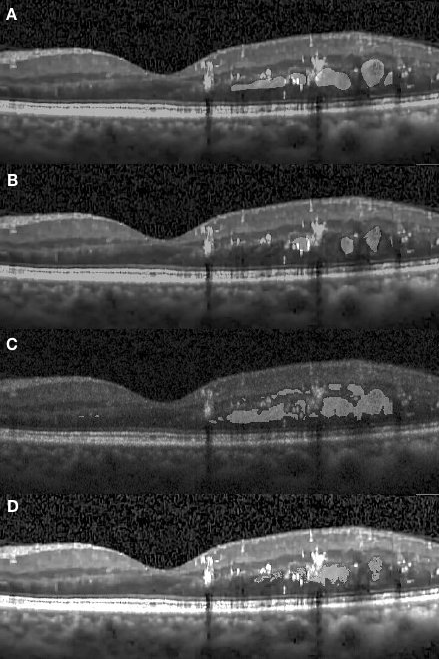}
\centering\includegraphics[width=.32\linewidth]{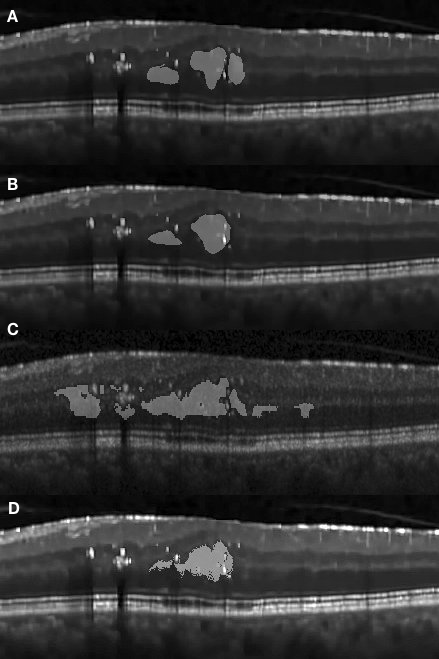}
\caption{A comparison between Chiu's work and the results obtained with our method. (A) and (B): annotations by two graders. (C): Segmentation results from Chiu's work. (D): our results. From the comparison, it can be seen that both sub-segmentation and over-segmentation are corrected in our results which have a higher agreement with manual annotation}
\label{Fig:9}
\end{figure*}

\begin{figure}[h!]
\centering\includegraphics[width=1\linewidth]{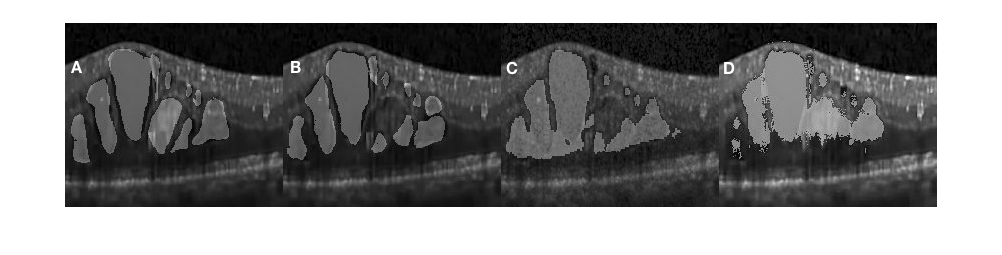}
\caption{A comparison between Chiu's work and the results obtained with our method. (A) and (B): annotations by two graders. (C): Segmentation results from Chiu's work. (D): our results. It is clearly shown that our results are adhering more consistently to the edema boundary of the manual segmentation.}
\label{Fig:10}
\end{figure}

Figure \ref{Fig:8} depicts the comparison of our method with the one reported in Chiu \textit{et al.} \cite{chiu2015kernel} for three segmentation attempts carried out complex retinopathy conditions. It is found that Chiu's method performs a slight over-segmentation near the bottom area. The segmentation results carried out with our method are closer to those which have been manually segmented. In Fig. \ref{Fig:9}, three images are shown which depict segmentation results on concrete and non-obvious fluid edema. The over-segmentation noted earlier can also be observed , as well as a false segmentation happening in the region on the left-hand side. Again, the results obtained with our method have a better agreement with those obtained via manual segmentation. Furthermore, our results are more adherent to target based on image content. This is also found in Fig. \ref{Fig:10} where additional results are presented from the segmentation of an example of fluid edema. Quantitatively, we compared the Dice overlap coefficient and $p$-value of Wilcoxon matched-pairs test for the DME region. Dice coefficients were calculated for all 10 patients and the process was repeated as per\cite{chiu2015kernel}. Specifically, the Dice coefficient was calculated based on all test images and the Wilcoxon matched-pairs test was calculated based on the mean Dice coefficient across all patients for our automated method and for the corresponding results from two graders. It was found that the Dice coefficient of our approach is $0.61\pm 0.21$ standing for mean and standard deviation which outperforms $0.51\pm 0.34$ reported in \cite{chiu2015kernel} (the higher, the better). Both methods are comparable with the Dice coefficient between manual graders ( $0.58\pm 0.32$). The $p$-value for our approach is $0.53$, which is also better than $0.43$ reported in Chiu \textit{et al.} \cite{chiu2015kernel} (where a coefficient value of $1$ indicates perfect agreement).

\section{Conclusions and future work}
\label{S:5}
In this paper, we presented a framework for segmenting CME regions in OCT retinal images by training fully convolutional neural networks (FCN) followed by post-precessing involving the fully connected CRFs (dense CRFs) method. The FCN model is an end-to-end network that takes advantage of the convolutional layers with skipping links to predict tissue classification at the pixel level. The skipping link allows the output from deeper convolutional layers to be fused with that from shallower convolutional layers. This feature compensates for the coarse prediction caused by down-sampling in the pooling layer and produces more refined results. We added the Dice layer to train the network where the overlap of predicted results and ground truth was taken into account during training process. We further fine-tuned the network using the pre-trained model and utilized data augmentation to compensate for the requirement of large amounts of training data. The dense CRFs are integrated to further refine the segmentation results. With dense CRFs, the recall rate of the boundaries was improved.

The segmentation results obtained with this framework demonstrates the potential abilities of end-to-end convolutional neural networks for OCT retinal image segmentation tasks. We emphasize that a key advantage of this method it's potential to cope with a variety of symptoms. The morphological properties of a certain retinopathy can be diverse and differently represented depending on different conditions. Conventional segmentation methods for OCT retinal image segmentation were often tailored for a specific task and their performance may vary depending on retinopathy condition. Conversely, convolutional neural networks can be trained to identify various retinopathy conditions, thus avoiding the requirement for a multitude of ad hoc rules. We intend to extend the application of this framework for retinal layer segmentation and segmentation of other diseases by training the network with larger training datasets via manual annotations of the symptoms of interest. It is expected that this learning based approach be adapted for more complex cases and can be more flexible than fixed mathematical model-based approaches. Therefore, this study is valuable for the final goal of developing a universal OCT segmentation approach.

The results from our experiments also identified some disadvantages of the framework. Although the network architecture involves the skip link to fuse the appearance information of shallow layers with the deeper layer outputs to make a more precise prediction, the structural character still presents some limitations in extracting accurate locations of small target regions. However, this disadvantage may be mitigated by modifying the network structure, namely by strengthening the end-to-end training.

In this paper, our experiment is focused on segmentation of cystoid macular edema. As future work, we intend to extend the method to several ophthalmic conditions, including deformed retinal layer segmentation caused by age-related macular degeneration, epiretinal membrane, macular hole and retinal detachment. We also intend to investigate the benefits of using more efficient network architectures to improve the segmentation accuracy. At the same time, we plan to incorporate super-voxel and recurrent neural networks as a post-processing step to enhance the inference efficiency with the aim of 3D segmentation which is more practical for quantitative analysis and clinical diagnosis.

\section*{Acknowledgements}
We appreciate the discussion and suggestion on framework architecture given by Jinchao Liu from Visionmetric Ltd, and the technical support during the development for system compatibility and hardware deployment from Nicholas French (SPS, Kent). We also appreciate the suggestions and comments about the manuscript from Sally Makin (SPS, Kent).





\bibliographystyle{model1-num-names}
\bibliography{sample.bib}







\end{document}